\documentclass[letterpaper, 10 pt, conference]{ieeeconf}  % Comment this line out if you need a4paper
\newcommand{\task}{T}

\newcommand{\symdp}{\mathcal{T}}

\newcommand{\macgyver}{\tilde{\symdp}}

\usepackage{mathtools}
\usepackage{graphicx}
\usepackage{booktabs}
\usepackage{algorithm}
\usepackage{algpseudocode}
\usepackage{adjustbox}
\usepackage{amssymb}
\usepackage{amsmath}
\usepackage{arydshln}
\usepackage{tabularray}
\usepackage{color, colortbl}
\usepackage[table]{xcolor} % For \rowcolor
\usepackage{hyperref}
\usepackage{array}
\usepackage{color}
\definecolor{ashgray}{rgb}{0.7,0.75,0.71}
\definecolor{babypink}{rgb}{0.96,0.76,0.76}
\definecolor{dollarbill}{rgb}{0.52, 0.73, 0.4}
\definecolor{pastelgreen}{rgb}{0.47, 0.87, 0.47}
\IEEEoverridecommandlockouts                              % This command is only needed if 
\title{\LARGE \bf
Curiosity-Driven Imagination: Discovering Plan Operators and Learning Associated Policies for Open-World Adaptation
}

\author{Pierrick Lorang$^{1,2}$, Hong Lu$^{1}$, and Matthias Scheutz$^{1}$%$<-this % stops a space
\thanks{$^{1}$Tufts University, 419 Boston Ave, Medford, 02155, MA, USA {\tt\small first.last@tufts.edu}}%
\thanks{$^{2}$AIT Austrian Institute of Technology GmbH, Center for Vision, Automation \& Control, Vienna, Austria {\tt\small first.last@ait.ac.at}}%
}

\begin{document}

\maketitle
\thispagestyle{empty}
\pagestyle{empty}

%%%%%%%%%%%%%%%%%%%%%%%%%%%%%%%%%%%%%%%%%%%%%%%%%%%%%%%%%%%%%%%%%%%%%%%%%%%%%%%%
\begin{abstract}

Adapting quickly to dynamic, uncertain environments—often called ``open worlds"—remains a major challenge in robotics. Traditional Task and Motion Planning (TAMP) approaches struggle to cope with unforeseen changes, are data-inefficient when adapting, and do not leverage world models during learning. We address this issue with a hybrid planning and learning system that integrates two models: a low-level neural network-based model that learns stochastic transitions and drives exploration via an Intrinsic Curiosity Module (ICM), and a high-level symbolic planning model that captures abstract transitions using operators, enabling the agent to plan in an ``imaginary" space and generate reward machines. Our evaluation in a robotic manipulation domain with sequential novelty injections demonstrates that our approach converges faster and outperforms state-of-the-art hybrid methods.

\end{abstract}

\section{Introduction}
\label{sec:intro}

In open-world scenarios~\cite{goeletal24aij}, unforeseen changes can invalidate an agent's pre-existing knowledge, requiring it to quickly adapt -- something purely symbolic approaches struggle to achieve~\cite{gizzi2019creative}. Previous work has sought to enhance existing Task and Motion Planning (TAMP) methods by integrating Reinforcement Learning (RL)~\cite{goel2022rapidlearn, peorl-Yang,Cheng_Xu_2023, Silver_Athalye_Tenenbaum_Lozano-Pérez_Kaelbling_2023, Kokel_Manoharan_Natarajan_Ravindran_Tadepalli_2021, Illanes_Yan_Icarte_McIlraith_2020, Acharya_Raza_Dourado_Velasquez_Song_2023, Lorang_Horvath_Kietreiber_Zips_Heitzinger_Scheutz_2024, Balloch_Lin_Wright_Peng_Hussain_Srinivas_Kim_Riedl_2023, Guan_Sreedharan_Kambhampati}, a hybrid approach that allows for handling novel situations by combining symbolic planning to reduce the complexity of decision-making with the flexibility of RL to learn new low-level policies. While such hybrid systems outperform end-to-end neural networks, they remain inefficient for real-world applications, often requiring millions of training interactions to adapt to even minor changes, which is due to their reliance on reward structures that are fixed, sparse, or require human input.
%, which fail to guide the learning algorithm effectively when environmental shifts alter the agent's original model of the world.

{\em Reward machines} that can ``densify'' reward signals by leveraging temporal logic inferences \cite{Gehring_Asai_Chitnis_Silver_Kaelbling_Sohrabi_Katz_2022, Icarte2020RewardME} have been proposed to address this limitation, but the challenge remains to determine an effective mechanism for inferring the logic that supports the reward machine and optimally guides RL within bi-level agents for rapid adaptation to novelties.

\begin{figure}[t]
  \centering
  \begin{minipage}[t]{0.48\textwidth}
    \includegraphics[width=0.98\textwidth]{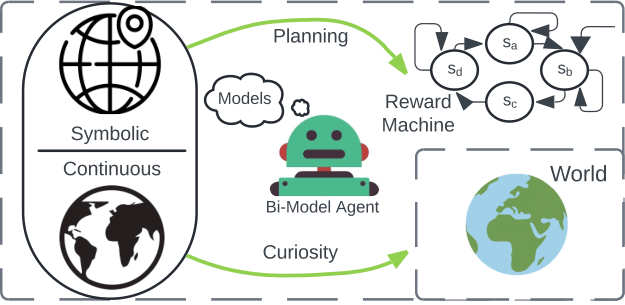}
    \caption{Curiosity-Driven Imagination: The agent learns a bi-level model of the environment—continuous (neural network) and symbolic (planning domain). The continuous component drives intrinsic curiosity, guiding the agent to unfamiliar states for symbolic abstraction, while the symbolic component constructs a reward machine based on hypothetical plans.}
    \label{fig:hybrid_planning_learning}
  \end{minipage}
  %\vspace{-0.75cm}
\end{figure}

We propose to enhance hybrid planning and learning systems with a dual-model learner  (Fig.~\ref{fig:hybrid_planning_learning}) which automates the creation of reward machines by improving exploration and exploitation in hybrid planning and RL, learning two transition models in parallel with policy training: a symbolic layer that abstracts high-level operators from environment interactions, and a stochastic neural network with an ``Intrinsic Curiosity Module'' (ICM)~\cite{pmlr-v70-pathak17a} that drives exploration towards novel transitions. By learning non-grounded symbolic transitions, the agent can reason in an ``imaginary'' space, inferring unseen transitions from analogous experiences and planning actions it has never encountered or executed before. These plans are then used to construct Linear Temporal Logic (LTL) reward machines~\cite{Camacho_Toro_Icarte_Klassen_Valenzano_McIlraith_2019}, exploiting the knowledge abstracted during exploration. The ICM further encourages exploration, aiding the discovery of new symbolic transitions and enhancing neuro-symbolic learning efficiency. Evaluated in the continuous robotics domain (Mujoco), our approach significantly outperforms traditional as well as the most recent RL and hybrid systems.

Our specific contributions include:
(1) A Numerical Planning Operator Learner that abstracts operators from symbolic transitions and generates LTL reward machines, enabling effective exploitation of symbolic information.
(2) An ICM-driven exploration protocol that guides the agent toward novel symbolic transitions, improving its adaptability in unfamiliar scenarios.

\section{Related Work}

% The design of integrated systems for addressing open-world novelties has emerged as a recent area of study~\cite{muhammad2021novelty, sarathy2021spotter, liu2023ai, goeletal24aij}. While hybrid methods offer potential solutions, they frequently encounter challenges related to data inefficiency and extended training periods, particularly in continuous environments. Recovering from planning failures through RL remains a challenge due to its high data demands, particularly when swift adaptation is required in novelty-rich scenarios. Recent research has primarily focused on settings where agents face informed, gradual, or plan-level changes~\cite{Icarte2020RewardME, seo2020trajectorywise, Nayyar_Verma_Srivastava_2022}. While some efforts aim for task novelty adaptation within a single trial, often at the cost of robustness~\cite{chen2022single}, others have explored human-assisted guidance~\cite{goel2022rapidlearn}. To address data efficiency, goal-oriented RL has been integrated into hybrid frameworks~\cite{xu2022framework, lorang2022speeding, lorangetal2024iros}, while hierarchical architectures~\cite{peorl-Yang, lorangetal24icra} have been employed to improve adaptability. However, these approaches typically rely on pre-defined, sparse, or human-aided reward systems.

The design of integrated systems to handle open-world novelties has gained attention recently~\cite{muhammad2021novelty, sarathy2021spotter, liu2023ai, goeletal24aij}. While hybrid methods show promise, they often struggle with data inefficiency and long training times, especially in continuous environments. Recovering from planning failures using RL remains challenging due to its high data demands, particularly in novelty-rich scenarios requiring quick adaptation. Most research has focused on informed, gradual, or plan-level changes~\cite{Icarte2020RewardME, seo2020trajectorywise, Nayyar_Verma_Srivastava_2022}, with some efforts targeting single-trial task novelty adaptation at the cost of robustness~\cite{chen2022single} or exploring human-guided approaches~\cite{goel2022rapidlearn}. 
%To improve data efficiency, goal-oriented RL has been integrated into hybrid frameworks~\cite{xu2022framework, lorang2022speeding, lorangetal2024iros}, or hierarchical architectures~\cite{peorl-Yang, lorangetal24icra} have emerged. However, these approaches typically rely on pre-defined, sparse, or human-aided reward systems.
To enhance data efficiency, goal-oriented RL has been integrated into hybrid frameworks~\cite{Xu_Fekri_2022, lorang2022speeding, lorangetal2024iros} or hierarchical architectures~\cite{peorl-Yang,  Lorang_Horvath_Kietreiber_Zips_Heitzinger_Scheutz_2024}. However, these methods typically rely on pre-defined, sparse, or human-aided rewards.

% Recent interest has shifted toward abstracting symbolic world models~\cite{Arora_Fiorino_Pellier_Métivier_Pesty_2018, Konidaris_Kaelbling_Lozano-Perez_2018, Loula_Allen_Silver_Tenenbaum_2020, Silver_Chitnis_Tenenbaum_Kaelbling_Lozano-Perez_2021, Chitnis_Silver_Tenenbaum_Lozano-Perez_Kaelbling_2022}, with promising results seen in learning neuro-symbolic models to improve adaptation in open-world environments~\cite{doncieux2020dream, Balloch_Lin_Wright_Peng_Hussain_Srinivas_Kim_Riedl_2023}. Simultaneously, curiosity-driven methods have demonstrated strong exploratory capabilities, allowing agents to abstract new knowledge~\cite{Chitnis_Silver_Tenenbaum_Kaelbling_Lozano-Pérez_2021, curtis2020flexible,sartor2023intrinsically}. The densification of reward systems through bi-level heuristics is also becoming a focus for guiding RL agents~\cite{Hu_Wang_Jia_Wang_Chen_Hao_Wu_Fan, Stadie_Zhang_Ba_2020, Gehring_Asai_Chitnis_Silver_Kaelbling_Sohrabi_Katz_2022}, particularly using reward machines based on temporal logic, which offer a robust solution for sparse feedback signals~\cite{li2017reinforcementlearningtemporallogic, Littman2017EnvironmentIndependentTS, Illanes_Yan_Icarte_McIlraith_2020, Hasanbeig_Jeppu_Abate_Melham_Kroening, Xu_Fekri_2022}.

Recent work has focused on abstracting symbolic world models~\cite{Arora_Fiorino_Pellier_Métivier_Pesty_2018, Konidaris_Kaelbling_Lozano-Perez_2018, Loula_Allen_Silver_Tenenbaum_2020, Silver_Chitnis_Tenenbaum_Kaelbling_Lozano-Perez_2021, Chitnis_Silver_Tenenbaum_Lozano-Perez_Kaelbling_2022}, showing promising results in neuro-symbolic models for better adaptation in open-world environments~\cite{doncieux2020dream, Balloch_Lin_Wright_Peng_Hussain_Srinivas_Kim_Riedl_2023}. Curiosity-driven methods also demonstrate strong exploratory capabilities, helping agents to abstract new knowledge~\cite{Chitnis_Silver_Tenenbaum_Kaelbling_Lozano-Pérez_2021, curtis2020flexible, sartor2023intrinsically}. Reward system densification through bi-level heuristics is emerging to guide RL agents~\cite{Hu_Wang_Jia_Wang_Chen_Hao_Wu_Fan, Stadie_Zhang_Ba_2020}, particularly with reward machines based on temporal logic, offering robust solutions for sparse feedback~\cite{li2017reinforcementlearningtemporallogic, Littman2017EnvironmentIndependentTS, Illanes_Yan_Icarte_McIlraith_2020, Hasanbeig_Jeppu_Abate_Melham_Kroening, Xu_Fekri_2022}.

In this work, we unify these various promising but disconnected approaches: leveraging symbolic abstraction to learn a symbolic representation of low-level transitions, employing stochastic models for curiosity-driven exploration, using neuro-symbolic world models to plan in imaginary spaces, and incorporating reward machines to densify feedback based on imaginary plans. This synergistic combination accelerates adaptation to abrupt, uninformed novelties in open-world environments without requiring human intervention.

\section{Preliminaries}

\begin{figure}[t]
  \centering
  \begin{minipage}[t]{0.47\textwidth}
    \includegraphics[width=0.98\textwidth]{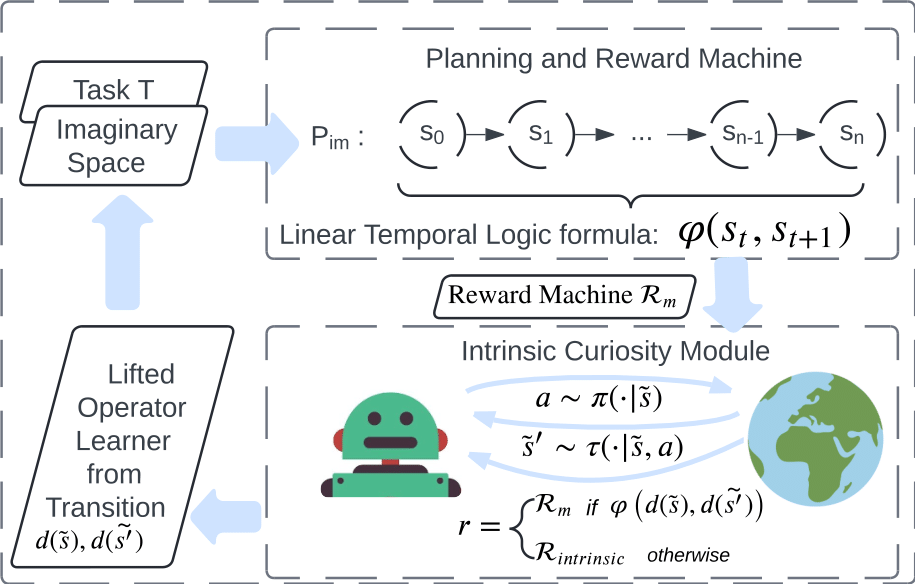}
    \caption{Operator Discovery: The agent identifies hypothetical lifted operators from symbolic transitions, which are employed for planning and generating an LTL formula to enhance the reward signal. The reward feedback comprises two components: a reward machine that activates when state transitions meet the LTL formula, and an intrinsic curiosity reward that encourages the agent to explore and learn new transitions.}
    \vspace{-0.5cm}
    \label{fig:bilevel}
  \end{minipage}
  %\vspace{-0.2cm}
\end{figure}

% \subsection{Symbolic Planning}
% \label{par:hybrid}

% Symbolic planning typically builds upon a domain rendered in a formal language such as PDDL \cite{mcdermott1998pddl}. Let $\sigma = \langle \mathcal{E}, \mathcal{F}, \mathcal{S}, \mathcal{O}\rangle$ be a domain description,

% where $\mathcal{E}=\left\{\varepsilon_{1}, \ldots, \varepsilon_{|\mathcal{E}|}\right\}$ is a set of entities within the environment. $\mathcal{F}=\left\{f_{1}(\odot), \ldots, f_{|\mathcal{F}|}(\odot)\right\}, \odot \subset \mathcal{E}$ is a set of boolean or numerical predicates. $\mathcal{S}=\left\{s_{1} \ldots, s_{|\mathcal{S}|}\right\}$ is the set of symbolic states in the environment. $\mathcal{O}$ represents the set of known action operators, defined as $\mathcal{O}=\left\{o_{1}, \ldots, o_{|\mathcal{O}|}\right\}$. Each operator $o_i$ is characterized by a set of preconditions and effects denoted $\psi_{i}, \omega_{i} \in \mathcal{F}$. The preconditions $\psi_{i}$ and effects $\omega_i$ of $o_{i}$ represents the predicates that must hold (or must not hold) before and after executing $o_{i}$, respectively. A planning task is typically described as a STRIPS task which we denote as $T=\left(\mathcal{E}, \mathcal{F}, \mathcal{O}, s_{0}, s_{g}\right)$, where $s_{0} \subset \mathcal{S}$ represents the set of initial states and $s_{g} \subset \mathcal{S}$ represents the set of desired goal states. The solution to this planning task $T$ is presented as an ordered sequence of operators, denoted as plan $\mathcal{P}=\left[o_{1}, \ldots o_{\mid \mathcal{P}]}\right]$.

\textbf{Symbolic Planning.} Symbolic planning typically relies on a domain described in a formal language such as the Planning Domain Description Language (PDDL) \cite{mcdermott1998pddl}. Let $\sigma = \langle \mathcal{E}, \mathcal{F}, \mathcal{S}, \mathcal{O}\rangle$ denote a domain description, where $\mathcal{E}=\left[\varepsilon_{1}, \ldots, \varepsilon_{|\mathcal{E}|}\right]$ represents a set of entities within the environment, and $\mathcal{F}=\left[f_{1}(\odot), \ldots, f_{|\mathcal{F}|}(\odot)\right]$, with $\odot \subset \mathcal{E}$, is a set of boolean or numerical predicates. The set $\mathcal{S}=\left[s_{1}, \ldots, s_{|\mathcal{S}|}\right]$ denotes the symbolic states in the environment, while $\mathcal{O}$ represents the set of known action operators, defined as $\mathcal{O}=\left[o_{1}, \ldots, o_{|\mathcal{O}|}\right]$. 
% Each operator $o_i \in \mathcal{O}$ is characterized by a set of preconditions and effects, denoted $\psi_{i}$ and $\omega_{i}$ respectively, where $\psi_{i}, \omega_{i} \in \mathcal{F}$. The preconditions $\psi_{i}$ are the predicates that must hold before executing $o_{i}$, and the effects $\omega_{i}$ are the predicates that must hold after executing $o_{i}$. A planning task is typically described as a STRIPS task, denoted $T=\left(\mathcal{E}, \mathcal{F}, \mathcal{O}, s_{0}, s_{g}\right)$, where $s_{0} \subset \mathcal{S}$ represents the set of initial states and $s_{g} \subset \mathcal{S}$ represents the set of goal states. The solution to this planning task $T$ is an ordered sequence of operators, represented as a plan $\mathcal{P}=\left[o_{1}, \ldots, o_{|\mathcal{P}|}\right]$.
Each operator \( o_i \in \mathcal{O} \) has preconditions and effects, denoted \( \psi_{i} \) and \( \omega_{i} \), where \( \psi_{i}, \omega_{i} \in \mathcal{F} \). The preconditions \( \psi_{i} \) must hold before executing \( o_{i} \), and the effects \( \omega_{i} \) hold after execution. A planning task is described as a STRIPS task \( T=\left(\mathcal{E}, \mathcal{F}, \mathcal{O}, s_{0}, s_{g}\right) \), where \( s_{0} \subset \mathcal{S} \) is the initial state and \( s_{g} \subset \mathcal{S} \) is the goal state. The solution is a sequence of operators, represented as a plan \( \mathcal{P}=\left[o_{1}, \ldots, o_{|\mathcal{P}|}\right] \).

% \subsection{Learning Operators from Symbolic Transitions}
% \label{sec:numerical_operators}

\textbf{Learning Operators from Symbolic Transitions.} The process of learning operators from symbolic transitions has been studied as a way to abstract operators by analyzing changes in effects and existing preconditions. A lifted operator, as opposed to a grounded one, define an action abstractly with variables that can later be instantiated with specific entities. Symbolic transitions are pairs of states $(s_t, s_{t+1})$, where $s_t \subseteq \mathcal{S}$ is the state of the environment before the action, and $s_{t+1} \subseteq \mathcal{S}$ is the state after the action. The goal is to extract the operator's preconditions and its effects that transform $s_t$ to $s_{t+1}$.

A lifted operator \( \mathcal{O}_{\text{lifted}} = \langle \psi, \omega \rangle \) is defined by preconditions \( \psi \subseteq \mathcal{F} \) and effects \( \omega \subseteq \mathcal{F} \). Changes in predicates \( f_i(\odot) \) between \( s_t \) and \( s_{t+1} \), whether boolean or numerical, determine the effects. Effects are predicates \( f_i(\odot) \) that change between \( s_t \) and \( s_{t+1} \), while preconditions are those that remain unchanged across multiple initial states where the same operator is applied. Numerical predicates may involve increments, decrements, or other transformations.

% \subsection{Reinforcement Learning and Intrinsic Curiosity Module}
% \label{sec:rl_icm}
\textbf{Reinforcement Learning and ICM.}
RL is a framework where an agent interacts with an environment, modeled as a Markov Decision Process (MDP) \( M = \langle \mathcal{\widetilde{S}}, \mathcal{A}, R, \tau, \gamma \rangle \). Here, \( \mathcal{\widetilde{S}} \) is the sub-symbolic state space, \( \mathcal{A} \) is the action space, \( \tau(\tilde{s}_{t+1} | \tilde{s}_t,a_t) \) is the transition probability, \( R(\tilde{s},a) \) is the reward function, and \( \gamma \in (0,1] \) is the discount factor. The goal is to learn a policy \( \pi(a|\tilde{s}) \) that maximizes expected cumulative reward, \( G_t = \sum_{k=0}^{\infty} \gamma^k R_{t+k} \). Sparse or noisy rewards make learning difficult, but curiosity-driven exploration, using intrinsic rewards, helps address this.

The Intrinsic Curiosity Module~\cite{pmlr-v70-pathak17a} encourages exploration by predicting the outcomes of the agent’s actions. It includes a forward model, predicting the next state, and an inverse model, predicting the action causing the transition. The intrinsic reward is based on the prediction error of the forward model: \( \mathcal{R}_{\text{intrinsic}} = ||\hat{s}_{t+1} - \tilde{s}_{t+1}||^2 \), where \( \hat{s}_{t+1} \) is the predicted state and \( \tilde{s}_{t+1} \) the actual state, motivating exploration of hard-to-predict states.

% \subsection{Hybrid Planning \& Learning for Novelty Accommodation}
% \subsection{Hybrid Plan \& Learn and Reward Machines}
% \label{par:hybrid}
\textbf{Hybrid Plan \& Learn and Reward Machines.}
In open-world scenarios, agents face unforeseen challenges. The Integrated Planning Task (IPT) framework integrates planning and learning~\cite{sarathy2021spotter}. An IPT \( \mathcal{T} = \langle T, M, d, e \rangle \) consists of a STRIPS task \( T \), an MDP \( M \), a detection function \( d: \widetilde{\mathcal{S}} \to \mathcal{S} \), and an execution function \( e: \mathcal{O} \to X_M \). Executors \( x = \langle I, \pi, \beta \rangle \in X_M \) include an initiation indicator \( I \), an RL policy \( \pi \), and a termination indicator \( \beta \). Solutions to an IPT are plans \( \mathcal{P} = [o_{1}, \ldots, o_{|\mathcal{P}|}] \), mapping operators to executors to reach the goal state \( \tilde{s} \subseteq s_g \).
For novel challenges, a ``Stretch-IPT'' \( \macgyver = \langle T, M', d, e' \rangle \) updates the MDP to \(M' = \langle \mathcal{\widetilde{S'}}, \mathcal{A}, R, \tau', \gamma \rangle \) with a new state space \( \mathcal{\widetilde{S'}} \) and transition function \( \tau' \). The adapting executor \( x_{adapting} \) is mapped to either modified or completely new operators. \( \macgyver \) is \textit{Solvable} if a path from \( s_0 \) to \( s_g \) exists.

Reward machines enhance hybrid systems by providing temporally sensitive rewards through Linear Temporal Logic (LTL)~\cite{li2017reinforcementlearningtemporallogic, Littman2017EnvironmentIndependentTS}. The reward function \( R(\tau) \) in an MDP is aligned with LTL specifications \( \varphi \), evaluated over trajectories \( \tau = (s_0, a_0, \ldots, s_T) \). Reward machines allow for more precise and adaptive reward signals that capture long-term goals and complex temporal dependencies, improving the agent's ability to handle dynamic and evolving environments.

\section{Method}

We define novelty as an unknown, and thus an unforeseen change in the dynamics $\tau$ or the state representation $\widetilde{\mathcal{S}}$ of the environment $M$. Our study focuses on scenarios where there is no symbolic solution because an operator is missing or fails due to these novelties, that is, the expected effects of an operator do not match the response of the environment. To adapt to such a scenario, our approach learns a new adapting executor $x_{\textit{adapting}}$ and modifies or discovers a new operator. We then map this executor to either an existing operator, which we adapt, or we learn a completely new operator, following \cite{lorangetal2024iros}.
%which requires the agent to learn a new executor that either adapts an existing operator or abstracts a completely new one. 
We call this challenge the ``adapting executor and operator discovery problem''. 
We aim to learn symbolic and continuous models of the environment alongside $x_{\textit{adapting}}$, leveraging bi-level models to guide and accelerate novelty accommodation.

\begin{algorithm}[t]
\footnotesize
\caption{~\textit{\textbf{Hybrid Planning \& Learning}~($\mathcal{\symdp}$)~}}\label{alg:Execution}
\begin{algorithmic}[1]
\State {$\mathcal{P} \leftarrow$ \textbf{Plan}($\task$)}\Comment{$\mathcal{P}=\langle o_1, o_2, ..., o_{|\mathcal{P}|}\rangle$}
\For {$o_i \in \mathcal{P}$}
\State \textit{success} $\leftarrow$  \textbf{Execute}($o_i$)
\If {$\neg$ \textit{success}} \Comment{Proceed to \textit{Adapting} and \textit{Discovery}}
\State $\macgyver, o_f, s_{f} \leftarrow$ \textbf{UpdateProblem}($\mathcal{\symdp}$) \Comment{Stretch IPT}
\State $o_{\textit{discovery}} \leftarrow$ \textit{\textbf{Curiosity-Driven Imagination}(~$\mathcal{\widetilde{T}}$, $o_f$, $s_{f}$)}
\State \textit{success} $\leftarrow$  \textbf{Execute}($o_{\textit{discovery}}$) \Comment{Execute Learned Executor} 
\If{$\neg$ success}
\State \Return \textit{false} \Comment{Abort if $\neg$\textit{success} after adapting an executor}
\EndIf
\EndIf
\EndFor
\State \Return success \Comment{Success of $\mathcal{P}$}

\end{algorithmic}
\end{algorithm}
%\vspace{-0.1cm}

%\subsection{Curiosity Based Exploration of Operators}
\textbf{Curiosity Based Exploration of Operators.}
% Our methodology incorporates a latent space curiosity-driven module, similar to ICM, that guides the agent toward un- or underexplored states using the reward function $\mathcal{R}_{\text{intrinsic}}$. This mechanism encourages exploration of continuous state spaces, focusing on areas that could reveal novel symbolic transitions not yet abstracted into the agent symbolic knowledge. For such an abstraction, we introduce a symbolic lifted operator learner that allows the agent to continuously abstract new symbolic transitions as PDDL operators during exploration. Each new transition is analyzed to infer a lifted operator $\mathcal{O}_{\text{lifted}} = \langle \psi, \omega \rangle$, where $\psi$ represents the preconditions and $\omega$ represents the effects. Depending on the nature of the predicates that change during the transition, Boolean or numerical, we abstract classical or numerical operators. The cost of each operator is set as being the number of steps the agent took to achieve the symbolic transition in the (low-level) environment. We merge operators according to traditional symbolic learning principles by removing inconsistent preconditions from operators with identical effects and selecting the operator with the lowest cost among merged operators.
Our methodology incorporates a curiosity-driven module, like ICM, guiding the agent toward un- or underexplored states via the reward function \( \mathcal{R}_{\text{intrinsic}} \). This encourages the exploration of continuous state spaces to reveal novel symbolic transitions. We introduce a symbolic lifted operator learner that enables the agent to abstract new symbolic transitions as PDDL operators. Each new transition is analyzed to infer a lifted operator \( \mathcal{O}_{\text{lifted}} = \langle \psi, \omega \rangle \), where \( \psi \) are preconditions and \( \omega \) are effects. Based on changing predicates (Boolean or numerical), we abstract classical or numerical operators. The cost of each operator equals the steps taken to achieve the symbolic transition in the environment. Operators are merged by removing inconsistent preconditions from those with identical effects and selecting the operator with the lowest cost.

%%%%%%%%%%%%%%%%%%%%%%%%%%%%%%%%%%%%%%%%%%%%%%%%%%%%%%%%%%%%

\begin{algorithm}[t]
\footnotesize
\centering
\caption{~\textit{\textbf{Curiosity-Driven Imagination}} ($\macgyver$, $o_f$, $s_{f}$) $\rightarrow$ $x_\textit{adapting}$}
\label{alg:learn_algorithm}
\begin{algorithmic}[1] %[1] enables line numbers

\Require Domain $\sigma_{im}$ populated with $\mathcal{E}$ and $\mathcal{F}$
\Require Planning Task $T = \langle \mathcal{E}, \mathcal{F}, \mathcal{O}_{im}, s_0, s_g \rangle$
\Require Curiosity-driven exploration module (e.g., ICM)
\Require Reinforcement learning algorithm $\mathbf{A}$
\Require Maximum number of episodes $N_{\textit{eps}}$ 
\Require Maximum number of steps per episodes $n_{\textit{steps}}$ 
\Require Logarithmic frequency of re-planning $f_p$ (episodes)
\Require{Success rate threshold $\eta$}

\State Initialize $\pi$ from $\mathbf{A}$
\State Constrain Lifted $o_f$ in $\mathcal{O}_{im}$ with grounded preconditions

\State $I \leftarrow \textit{test(}d(\tilde{s}) \textbf{ is } s_f)$ \Comment{Initiation indicator}
\State $\beta \leftarrow \textit{test(}d(\tilde{s}) \in \mathcal{G}_{\mathcal{P}})$ \Comment{Termination indicator}
\State $T_k \leftarrow \emptyset$ \Comment{Set of known transitions}

\For{episode $\in$ \textit{$N_{eps}$}}
\State $\tilde{s} \leftarrow \textbf{Sample\_state(}M', I)$ \Comment{Reset the MDP s.t. $I(\tilde{s}) \textbf{ is }1$}

\If{$f_p(episode) = 1$} \Comment{PRM}
\State $\mathcal{P}_{im} \leftarrow \textbf{Plan}(\sigma_{im}, T)$ \Comment{$\mathcal{P}_{im}=\langle o_1, o_2, \dots, o_{\mid \mathcal{P}_{im} \mid} \rangle$}
\For{each symbolic transition $(s_t, s_{t+1})$ in $\mathcal{P}_{im}$}
    \State $\varphi(s_t, s_{t+1}) \leftarrow$ \textbf{GenerateLTLFormula}$(s_t, s_{t+1})$ 
\EndFor

\State $\mathcal{R}_{\text{m}} \leftarrow$ \textbf{CreateRewardMachine}$(\varphi)$

\EndIf

\State done $\leftarrow$ \textit{false}
\While{$\neg$done}
\State $s \leftarrow d(\tilde{s})$
\State $a \sim \pi(\cdot | \tilde{s})$
\State $\tilde{s}' \sim \tau(\cdot | \tilde{s},a)$ \Comment{Environment step}
\State $s' \leftarrow d(\tilde{s}')$
\If{$(s,s') \notin T_k$}
\State $T_k \leftarrow T_k \cup \{(s,s')\}$
\EndIf
\State $\mathcal{R}_{\text{intrinsic}} \leftarrow$ \textbf{CuriosityReward}$(\tilde{s}')$ \Comment{ICM reward}
\State $r \leftarrow$ \textbf{ComputeReward}($\mathcal{R}_{\text{m}}$, $\mathcal{R}_{\text{intrinsic}})$

\State $\pi \leftarrow $\textit{\textbf{$\mathbf{A}_{\textit{Update}}$($\tilde{s}$,~$a$,~$\tilde{s}'$,~$r$)}}

\If{$(\beta(\tilde{s}')\textbf{ is }1) \lor  \textit{reached}(n_{\textit{steps}})$} \Comment{Termination}
\State done $\leftarrow$ \textit{true}
\EndIf
\State $\tilde{s} \leftarrow \tilde{s}'$
\EndWhile

\If{ $\textit{success}(\pi, \tilde{\mathcal{T}}) > \eta$}
%\State $\pi^{c}_{x'_i} \!\leftarrow \!\pi$, $x'_i\! \leftarrow\! \langle \tilde{S}'_0,\pi^{c}_{x'_i},\beta_{x'_i} \rangle$
\State $x_{\textit{adapting}}\! \leftarrow\! \langle I,\pi,\beta \rangle$ \Comment{Convergence check}
\State $o_{\textit{discovery}} \leftarrow$ \textbf{Abstract}($x_{\textit{adapting}}$)
\Return $o_{\textit{discovery}}$
\EndIf

\State $\mathcal{O}_{im} \leftarrow$ \textbf{OperatorLearner($\mathcal{O}_{im}$, $T_k$)} \Comment{Update $\sigma_{im}$}
\State \textbf{UpdateCost}($\langle o_1, o_2, \dots, o_{\mid \mathcal{P}_{im} \mid} \rangle$), \textbf{UpdateICM}
%\State \textbf{UpdateICM}

\EndFor
% \Return \textit{failure}
\State $x_{\textit{adapting}}\! \leftarrow\! \langle I,\pi,\beta \rangle$
\State $o_{\textit{discovery}} \leftarrow$ \textbf{Abstract}($x_{\textit{adapting}}$)
\State \textbf{return} $o_{\textit{discovery}}$
\end{algorithmic}
\label{alg:adapt}
% \vspace{-3em}
\end{algorithm}

%\subsection{Constraining the Symbolic Imaginary Space}
\textbf{Constraining the Symbolic Imaginary Space.}
%Learning lifted operators helps reasoning in an imaginary space, as it enables inferring unseen symbolic transitions from analogical experiences. For example, the agent might learn a lifted operator \textit{pick(?object)} from picking an apple and then generate a plan to \textit{pick(orange)} without having ever picked up an orange or knowing how to execute that action. This leads us to refer to the symbolic domain being learned during exploration as ``imaginary'' (or hypothetical), since it allows the agent to reason about possible symbolic paths to the goal without prior experience or knowledge of successful execution. Formally, the imaginary domain is then $\sigma_{im} = \langle \mathcal{E}, \mathcal{F}, \mathcal{S}_{im}, \mathcal{O}_{im} \rangle$. This domain reuses the entities in $\mathcal{E}$ and the predicates in $\mathcal{F}$, yet consists of imaginary states and operators: $\mathcal{S}_{im}$ and $\mathcal{O}_{im}$. 
Learning lifted operators enables reasoning in an imaginary space, inferring unseen symbolic transitions from analogical experiences. For instance, an agent might learn a lifted operator \textit{pick(?object)} from picking an apple and then plan to \textit{pick(orange)} without  having ever picked up an orange or knowing how to execute that action. This leads us to refer to the symbolic domain being learned during exploration as ``imaginary'' (or hypothetical), since it allows the agent to reason about possible symbolic paths to the goal without prior experience or knowledge of successful execution. Formally, the imaginary domain is then $\sigma_{im} = \langle \mathcal{E}, \mathcal{F}, \mathcal{S}_{im}, \mathcal{O}_{im} \rangle$. This domain reuses the entities in $\mathcal{E}$ and the predicates in $\mathcal{F}$, yet consists of imaginary states and operators: $\mathcal{S}_{im}$ and $\mathcal{O}_{im}$.

However, learning lifted operators can be challenging when dealing with environmental novelties, as lifted transitions in the imaginary space may not hold in the actual environment, potentially leading to flawed planning. To mitigate this, our agent reconstructs the symbolic imaginary space for each new novelty. Importantly, the agent uses information from failed operators \( o_f \in \mathcal{O} \)—those with at least one mapped executor that can be tested, unlike operators in \( \mathcal{O}_{im} \)—to refine the imaginary space. For example, if the agent fails to execute the grounded operator \texttt{reach(table)} due to an obstructing door (i.e., \texttt{open(door)} is \textit{false}), we constrain the lifted \texttt{reach($\odot$)} transition with the precondition \texttt{not(= ?location table)} in the updated imaginary space. The agent may not initially know that the failure is due to the door; the transition \texttt{reach($\odot$)} might still work for \texttt{reach(chair)} or \texttt{reach(kitchen)} despite detecting \texttt{not(open(door))}. The agent must re-experience \texttt{reach(table)} in the modified environment to identify the necessary grounded predicates, such as \texttt{open(door)}, before accurately abstracting the operator. In this example, the lifted operator is then constrained as:
\[
\textit{reach(?location)}\left\{
\begin{aligned}
    &\textit{pre: not(?location table)}, \\
    &\textit{eff: (at ?location)}
\end{aligned}
\right\}
\]
and after experiencing that \textit{open(door)} is a necessary precondition in state $s_t$ for the effects of \textit{reach(table)} to be accessible in the subsequent state $s_{t+1}$, the agent can abstract the missing transition as a new grounded operator in $\mathcal{O}_{im}$:
\[
\textit{reach(table)}\left\{\textit{pre: (open door)}, \textit{eff: (at table)}\right\}
\]

This process ensures that the agent re-grounds operators that failed due to novelty, updating their preconditions and effects based on the altered environment before abstracting them again. A grounded representation of a symbolic transition that has been successfully executed is considered valid and can be reliably used for future planning. This protocol guarantees that critical transitions, essential for avoiding planning impasses, are accurately represented. Additionally, since this guarantee does not apply to all symbolic transitions that are lifted, we modify the cost of operators in the plan used to generate the reward machine for RL to avoid training using a dead end reward machine. Formally in our current methodology, we average the episodic sum of reward over $T_e$ episodes. When $T_e$ improves we decrease the cost of the operators in $\mathcal{P}_{im}$ by a factor of $\theta$. If the agent fails to improve, we increase the costs of the operators in $\mathcal{P}_{im}$.
Given:
%\begin{equation}
$\bar{R}_{T_e} = \frac{1}{T_e} \sum_{\text{episode}}^{T_e} R_{\text{episode}}$.
%\end{equation}
If $C_i=\textit{cost}(o_i)$, we get, $\forall o_i \in \mathcal{P}_{im}$,
\begin{equation}
C_i^{\text{new}} = \begin{cases} 
C_i \cdot \frac{1}{\theta} & \text{if } \bar{R}_{T_e} > \bar{R}_{T_e-1} \\
C_i \cdot \theta & \text{if } \bar{R}_{T_e} \leq \bar{R}_{T_e-1}.
\end{cases}
\end{equation}
%Alternatively, gradient descent could be explored in future works to fine-tune operator costs over episodes.

%\subsection{Symbolic Exploitation through Reward Machines}
\textbf{Symbolic Exploitation through Reward Machines.}
To exploit the learned symbolic information into the reinforcement learning protocol, we use a Planning and Reward Machine system (PRM). The PRM system generates reward machines from the symbolic plans derived from our planning framework applied in the imaginary domain $\sigma_{im}$. Given a domain $\sigma_{im}$ that gets updated over exploration and a planning task $T = \left(\mathcal{E}, \mathcal{F}, \mathcal{O}_{im}, s_0, s_g\right)$, we first obtain a sequence of operators $\mathcal{P}_{im} = \left[o_1, o_2, \ldots, o_{\mid \mathcal{P}\mid}\right]$ that represents an imaginary plan to achieve the goal state $s_g$. This plan is populated in the symbolic space to determine transitions between imaginary states, generating symbolic transitions $(s_t, s_{t+1})$. To guide reinforcement learning, we create a reward machine $\mathcal{R}_{\text{m}}$ that encodes these transitions using LTL (see~\cite{Camacho_Toro_Icarte_Klassen_Valenzano_McIlraith_2019} for details). Specifically, we define $\mathcal{R}_{\text{m}}$ using an LTL formula $\varphi$ that captures the desired behavior based on the plan $\mathcal{P}_{im}$. In particular, $\varphi$ may include temporal operators such as \textit{G} (globally), \textit{X} (next) and \textit{E} (eventually) to specify that certain conditions must hold in all states or eventually be achieved. Take as example an imaginary plan $\mathcal{P}_{im} = \langle (\texttt{turn\_on\_light}), (\texttt{pick\_object}), (\texttt{place\_object}) \rangle$. In this plan, a robotic arm must first turn on the light before performing the pick-and-place operation. The state sequence depicting the plan would then be:

% \begin{itemize}[itemsep=0pt]
%     \item
%     \{$\neg$\texttt{light\_on}, $\neg$\texttt{picked}, $\neg$\texttt{placed}\} \\
%     \item 
%     \{ \texttt{light\_on}, $\neg$\texttt{picked}, $\neg$\texttt{placed}\} \\
%     \item
%     \{ \texttt{light\_on}, \texttt{picked}, $\neg$\texttt{placed}\} \\
%     \item
%     \{ \texttt{light\_on}, $\neg$\texttt{picked}, \texttt{placed}\} \\
% \end{itemize}
\begin{align*}
\begin{tabular}{|c|l|}
\hline
\textbf{State Number} & \textbf{State Description} \\
\hline
0 & \{$\neg$\texttt{light\_on}, $\neg$\texttt{picked}, $\neg$\texttt{placed}\} \\
1 & \{ \texttt{light\_on}, $\neg$\texttt{picked}, $\neg$\texttt{placed}\} \\
2 & \{ \texttt{light\_on}, \texttt{picked}, $\neg$\texttt{placed}\} \\
3 & \{ \texttt{light\_on}, $\neg$\texttt{picked}, \texttt{placed}\} \\
\hline
\end{tabular}
\end{align*}

The LTL formula:
\begin{align*}
\varphi = &\textbf{E} (\texttt{light\_on})\land \textbf{X} (\textbf{E} (\texttt{picked})) \\
    &\land \textbf{X} (\textbf{E} (\texttt{placed} \land \neg \texttt{picked}))
\end{align*}
captures the intended behavior of $\mathcal{P}_{im}$. The agent should eventually turn on the light. Once the light is on, the agent must next eventually pick the object and then next eventually place it at the destination, completing the task in sequence. \textbf{G} would be used only for static positive predicates across transitions.

% The LTL formula:
% \begin{align*}
% \varphi = \textbf{G} \big( &(\neg \texttt{light\_on} \rightarrow \textbf{X} ( \textbf{F} (\texttt{light\_on})) \\
%     &\land (\texttt{light\_on} \rightarrow \textbf{X} ( \textbf{F} (\texttt{picked})) \\
%     &\land (\texttt{picked} \rightarrow \textbf{X} ( \textbf{F} (\texttt{placed} \land \neg \texttt{picked})) \big) \\
% \end{align*}
% captures the intended behavior of $\mathcal{P}_{im}$. It ensures that, at any point, if the light is off, it will eventually be turned on in the next state. Once the light is on, the agent must next pick the object and then place it in the destination, completing the task in sequence. In such example, \textbf{F} is the basic operator computed from the failed operator expected effects, i.e., the goals of the learning. \textbf{G}

The reward system we use to merge ICM and PRM can be expressed as:
\vspace{-0.25cm}
\begin{align*}
r(s_t, s_{t+1}) = \begin{cases} \mathcal{R}_{\text{m}} & \text{if } \varphi(s_t, s_{t+1}) \text{ is satisfied} \\ \mathcal{R}_{\text{intrinsic}} & \text{otherwise.} \end{cases}\\
\vspace{-1cm}
\end{align*}
%For instance, a predicate remaining \textit{True} starting from the $k^{th}$ step in the plan would be expressed as $G(k \leq t \rightarrow p_k)$, where $p_k$ is the predicate, indicating that $p_k$ must remain true from step $k$ onward. 
By incorporating such temporal logic constraints, the reward machine ensures that the agent is consistently guided by symbolic information across its exploration, helping to align the learned policy with the desired symbolic transitions in the imaginary domain. This mechanism allows the agent to both pursue intrinsic curiosity and adhere to goal-driven symbolic plans, leading to more efficient and structured learning in complex environments. The reward machine $\mathcal{R}_{\text{m}}$ thus encodes the progression through the symbolic space towards the goal, providing a reward signal that drives the reinforcement learning agent to (1) reach checkpoints in the imaginary plan $\mathcal{P}_{im}$ and (2) become curious about unfamiliar states.

%\subsection{Algorithmic Description}
\textbf{Algorithmic Description.}
%\textbf{Hybrid Planning \& Learning.} 
The high level execution protocol of the Hybrid Planning and Learning framework is shown in Alg.~\ref{alg:Execution} (based on \cite{lorangetal2024iros}). The agent iterates over the plan (lines 2-12) until a novelty in the environment modifies $M$ to $M'$, causing an execution impasse. To adapt to such novelty, the agent updates the IPT as a stretch IPT using the modified environment identifies the failure information (line 5). Afterward the agent enters our Curiosity-Driven Imagination protocol (line 6) to quickly discover a new operator ($o_{\textit{discovery}}$).
Upon successful execution of the discovery operator, the agent can continue with the rest of the plan $\mathcal{P}$; otherwise returns \textit{false}. For more details about Alg.~\ref{alg:Execution}, see \cite{lorangetal2024iros}.

%\textbf{Bi-Level Curiosity.} 
Our Bi-Level Curiosity algorithm, as illustrated in Fig.\ref{fig:bilevel}, is given in Alg.~\ref{alg:adapt}. The agent begins by initializing an RL policy (line 1) and an imaginary domain $\sigma_{im}$, constrained by failed operators (line 2). It then sets up the initiation function $I$, termination function $\beta$, and known transitions $T_k$ (lines 3–5). The agent proceeds with $N_{eps}$ episodes where it (1) plans in the imaginary space and generates a reward machine via PRM (lines 8–14), (2) explores the environment, calculates the ICM curiosity reward and updates the policy (lines 15–30), and (3) learns symbolic operators from new transitions $T_k$, updating $\sigma_{im}$ and the ICM network (line 36). If convergence is reached, it abstracts $x_{\textit{adapting}}$ as the discovered operator $o_{\textit{discovery}}$ and returns it (lines 33-34); otherwise, it returns it line 41.\footnote{Code available via this \href{https://github.com/lorangpi/PRM}{link}.}

%%%%%%%%%%%%%%%%%%%%%%%% ALGO %%%%%%%%%%%%%%%%%%%%%%%%%%%%%%%%%%%%%%%%%%%%%%%%
%
%%%%%%%%%%%%%%%%%%%%%%%%%%%%%%%%%%%%%%%%%%%%%%%%%%%%%%%%%%%%%%%%%%%%%%%%%%%%%%%%%%%%%%%%%%%%%%%%%%%%%%%%%%%%%%%%%%%%%%%%%%%%%%%

% -----------------------------------------------------------------------

% -----------------------------------------------------------------------
\section{Experiment}
\label{sec:experiments}

% \subsection{Environment}
% \label{sec:environments}

\begin{figure}[t]
  \centering
  \begin{minipage}[t]{0.47\textwidth}
    \includegraphics[width=0.98\textwidth]{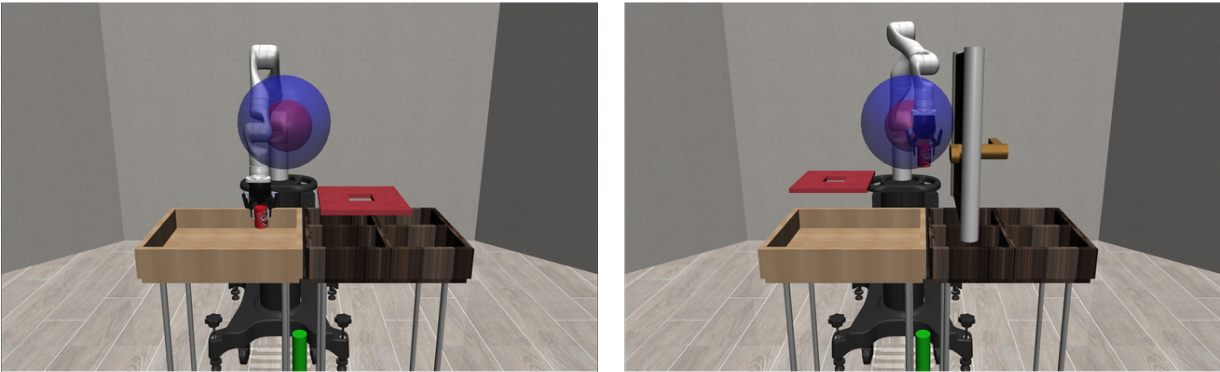}
    \caption{\textit{\textbf{Left:}} The original Pick\&Place task, where the agent must place a can in the drop-off bin. \textit{\textbf{Right:}} The \textit{Locked Door} novelty, where a door blocks access to the drop-off bin. The agent must first unlock the door via a proximity sensor (blue ball) before pushing it open. The red ball marks the light-switch location.}
    \label{fig:Fig1}
  \end{minipage}
  \vspace{-0.7cm}
\end{figure}

\textbf{Environment.} We evaluate our approach in RoboSuite~\cite{robosuite2020} {\em Pick and Place Can} task where a robotic arm must place a can in a bin. This task, despite being simple, poses challenges for RL agents~\cite{pmlr-v87-fan18a}. The observation space includes the robot's joint states and object positions/orientations, while the action space controls the end-effector's 3D displacement and gripper aperture.

\textbf{Experimental Scenario.} We replicate the setup from \cite{lorangetal2024iros} to compare the proposed algorithm with recent successful hybrid systems. We introduce novelties that obstruct essential operators such as \texttt{Pick($\cdot$)}, \texttt{Reach($\cdot$)}, and \texttt{Place($\cdot$)}, which fall into two categories: \textit{Shift}, where objects are displaced but remain detectable, and \textit{Disruption}, which introduces binary blocking mechanisms (see \cite{lorangetal2024iros} for details). These novelties are introduced sequentially, without overlap, and each agent is reset to its base performance (near 100\% success rate on the base task) between each novelty. As a result, the order of introduction does not affect the outcome.
\begin{figure*}[t]
  \centering
  \includegraphics[width=\textwidth]{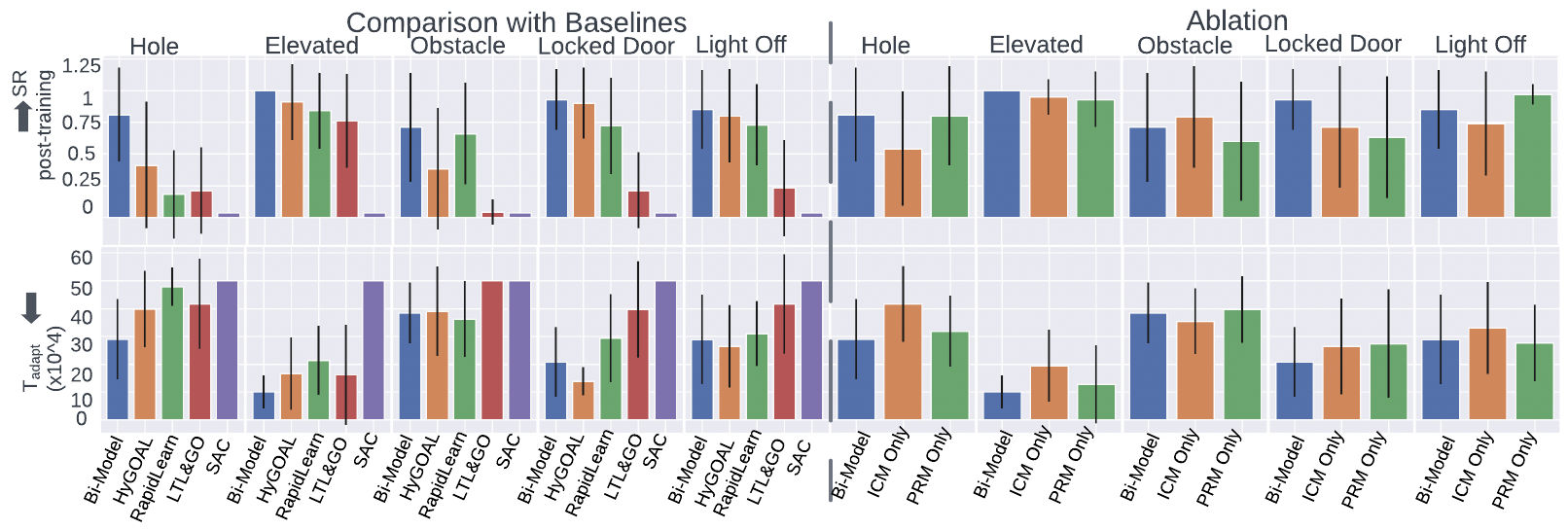}
  \vspace{-0.8cm}
  \caption{Experimental Results. The upward arrow next to SR indicates that higher is better whereas the downward arrow next to $T_{adapt}$ indicates that lower is better. \textit{Left}: Main results compared to the baselines. \textit{Right}: Ablation studies results with/without ICM/PRM. See text for more details.}
  %\vspace{-0.8cm}
  \label{fig:results}
\end{figure*}

\textbf{Detection function and Baselines.} The detection function $d$ maps sensor observations to symbolic states, using Boolean and numerical predicates. Our approach, \textit{Bi-Model}, is compared against several hybrid symbolic learning methods like \textit{HyGOAL}~\cite{lorangetal2024iros} (from which we reuse the experimental setup and the baselines results), \textit{RapidLearn}~\cite{goel2022rapidlearn}, and \textit{LTL\&GO} (GO for low-level Goal-Oriented learning)~\cite{Xu_Fekri_2022}, with Soft Actor-Critic (SAC) and Hindsight Experience Replay (HER) used as RL algorithms. Furthermore, we conducted ablation experiments without PRM or ICM to determine the contributions of each to the overall performance gains. 

\textbf{Evaluation and Metrics.} After each novelty injection, agents were trained until they achieved a success rate $80\%$ or completed $500,000$ steps. Each episode in RoboSuite consisted of up to $1,000$ interactions, with results averaged over $10$ seeds per agent. Performance was measured in every $20,000$ step by running $20$ evaluation episodes and calculating the mean success rate. Consistent RL hyperparameters were applied in all novelties for all baselines. We used $T_e=5$ and $\theta=1.05$ as parameters in our approach. The key metrics were $\text{T}_{\text{adapt}}$ (time steps to convergence) and $\text{SR}_{\text{post-training}}$ (success rate at asymptotic convergence).

\section{Results}
\label{sec:results}

%\vspace{-0.2em}
\textbf{Main Experiments.} Fig~\ref{fig:results} summarizes all agents' performance. Pure RL, along with re-planning (not listed in the table), yielded zero success across all novelties within $500,000$ steps, highlighting the inefficiency of non-hybrid RL in handling novelties, as also supported by RoboSuite Benchmark, and underscoring intrinsic RL limitations even in basic robotic tasks. Conversely, hybrid approaches demonstrated adaptability to five novelties, effectively managing significant task alterations by utilizing domain abstraction for targeted updates where novelties impact the agent's knowledge base.

Our method, \textit{Bi-Model}, demonstrated superior sample efficiency compared to other hybrid approaches, outperforming in all five scenarios in terms of asymptotic success rate and four of them in terms of adaptation time. In uninformed scenarios, where human guidance is not available, most baselines struggle to accelerate adaptation. The exception is \textit{HyGOAL}, which still performed equally well and faster in the \textit{Locked} novelty scenario. This advantage is due to \textit{HyGOAL} not requiring human guidance and its flexibility to not be restricted to experiencing specific state transitions for planning, unlike \textit{Bi-Model}. While \textit{HyGOAL} benefits from this flexibility, especially in scenarios requiring rare transitions that \textit{Bi-Model} needs to encounter to plan effectively in the imaginary space, \textit{Bi-Model} overall provides better performance. \textit{Bi-Model} leverages curiosity to gather symbolic transitions and efficiently exploits high-level operators in the imaginary space to find robust paths and densify the reward, which in turn leads to faster and better performance. 

\textbf{Ablation Experiments.} The ablation results of the \textit{Bi-Model} with and without PRM/ICM are also shown in \ref{fig:results}. The experiments confirm that ICM and PRM facilitate rapid and efficient adaptation in open-world environments. ICM drives exploration of unfamiliar states, often leading the agent toward task completion, while PRM densifies the rewards by abstracting symbolic transitions toward the goal. Although both methods guide learning effectively, it is their combination in \textit{Bi-Model} that leads to superior performance.

Individually, ICM and PRM have notable limitations. ICM strongly relies on its ability to quickly learn an accurate stochastic model of the transitions. For instance, in the \textit{Hole} scenario, the shape is difficult to model quickly and its location is slightly shifted randomly, making the learned model noisy. Furthermore, ICM may overly focus on curiosity, leading to exploration of novel but irrelevant states, especially in highly variable or abrupt scenarios. In \textit{LightOff}, the LiDAR sensor provides flat feedback until the light is restored, causing excessive exploration due to ICM later. PRM, on the other hand, may struggle with sparse or misleading symbolic signals, such as in the \textit{Obstacle} scenario, where detection is not dense enough to abstract useful operators for planning. Similarly, in the \textit{Locked} scenario, PRM struggles to plan for rare transitions until they are experimentally encountered and modeled.
The \textit{Bi-Model} integrates both approaches, overcoming these challenges and delivering a more robust performance across a range of environments.

\section{Discussion}
\label{sec:discussion}

Learning models of the environment is crucial for an agent to grasp the underlying dynamics and physics of its surroundings. Just as humans rely on cognitive models to both spark curiosity when unexpected phenomena occur and to predict the outcomes of their actions, artificial agents can benefit from similar mechanisms. These models allow us to anticipate, plan, and imagine pathways to achieve specific goals, often adjusting in real time to new information. Emulating human-like cognitive process within a neurosymbolic architecture enables artificial agents to navigate complex, dynamic environments with greater speed and precision. By integrating bi-level models—--symbolic planning at a high level and curiosity-driven exploration at a low level--—our approach allows the agent to not only generate effective plans but also explore novel states, accelerating adaptation to environmental changes.

Despite outperforming state-of-the-art algorithms, our approach faces scalability challenges. As \textit{Bi-Model} continuously abstracts symbolic operators, the domain grows, increasing planning costs and slowing learning. To address this, we reduced re-planning frequency in the imaginary space over time, though this only partially mitigates the issue. Future work should focus on optimizing operator quality over quantity. Our method reached convergence 20\% faster than \textit{HyGOAL}, the latest baseline. This difference stems from two key factors: faster step-wise convergence in novelty scenarios and \textit{HyGOAL}'s reliance on hindsight experience replay and planning towards an expanding set of goals, both of which are computationally expensive. However, our agent still requires tens of thousands of training steps to adapt to environmental changes, highlighting the need for further advancements and novel techniques.

Another limitation is the reliance on a reliable detection function and the assumption that lifted operators generalize across groundings (e.g., \textit{pick(apple)} to \textit{pick(orange)}). To address this, the agent should actively test abstracted operators in the real environment. Currently, failed operators in $\mathcal{O}$ are referenced for testing, but abstracted transitions in $\mathcal{O}_{im}$ are not verified. Future research should focus on validating these transitions in real-world settings to align symbolic knowledge with practical execution.

\section{Conclusion}

We proposed a hybrid bi-level model-driven learning approach incorporating ``intrinsic curiosity''-driven exploration with planning ``reward machines'' to address the challenge of coping with open-world novelties in dynamic environments.  Our method showed superior performance in pick-and-place tasks with introduced novelties not only to traditional but also to state-of-the-art hybrid methods, proving to be more sample efficient and adaptable to unknown contexts.  As AI systems increasingly face unpredictable settings, neuro-symbolic architectures like the one proposed offer a promising path toward achieving human-level adaptability. Future work will focus on enhancing the scalability and flexibility of this approach to even more complex real-world tasks.

\section*{Acknowledgments}

This work was in part funded by ONR grant \#N00014-24-1-2024.

\bibliographystyle{IEEEtran}
%\bibliographystyle{unsrtnat}
% \balance
\bibliography{main}

\clearpage

\end{document}